\documentclass[10pt,twocolumn,letterpaper]{article}

\usepackage{cvpr}
\usepackage{times}
\usepackage{epsfig}
\usepackage{graphicx}
\usepackage{subfigure}
\usepackage{amsmath}
\usepackage{amssymb}
\usepackage{multirow}
\usepackage{algorithm}
\usepackage{algorithmic}
\usepackage{amssymb}
\usepackage{eqparbox}

\usepackage{multicol}

\usepackage{makecell}
\usepackage[numbers]{natbib}

\newcommand{\delete}[1]{\textcolor[rgb]{0.545,0.545,0}{}}

\usepackage[pagebackref=true,breaklinks=true,letterpaper=true,colorlinks,bookmarks=false]{hyperref}

\cvprfinalcopy 

\def\cvprPaperID{1696} 

\begin{document}

\title{Memory Enhanced Global-Local Aggregation for Video Object Detection}

\author{Yihong Chen\textsuperscript{1,4}$\thanks{This work is done when Yihong Chen is an intern at Microsoft Research Asia.}$ \quad Yue Cao\textsuperscript{3} \quad Han Hu\textsuperscript{3} \quad Liwei Wang\textsuperscript{1,2}\\
\textsuperscript{1}Center of Data Science, Peking University \\
\textsuperscript{2}Key Laboratory of Machine Perception, MOE, School of EECS, Peking University \\ 
\textsuperscript{3}Microsoft Research Asia \quad \textsuperscript{4}Zhejiang Lab\\
{\tt\small chenyihong@pku.edu.cn \quad \{yuecao,hanhu\}@microsoft.com \quad wanglw@cis.pku.edu.cn}
}

\maketitle

\begin{abstract}
   How do humans recognize an object in a piece of video? Due to the deteriorated quality of single frame, it may be hard for people to identify an occluded object in this frame by just utilizing information within one image. We argue that there are two important cues for humans to recognize objects in videos: the global semantic information and the local localization information. Recently, plenty of methods adopt the self-attention mechanisms to enhance the features in key frame with either global semantic information or local localization information. In this paper we introduce memory enhanced global-local aggregation (MEGA) network, which is among the first trials that takes full consideration of both global and local information. Furthermore, empowered by a novel and carefully-designed Long Range Memory (LRM) module, our proposed MEGA could enable the key frame to get access to much more content than any previous methods. Enhanced by these two sources of information, our method achieves state-of-the-art performance on ImageNet VID dataset. Code is available at \url{https://github.com/Scalsol/mega.pytorch}.
\end{abstract}

\section{Introduction}\label{sec:intro}

\begin{figure}[t]
\begin{center}
\subfigure[video object detection with full connection.]{
    \begin{minipage}[t]{0.9\linewidth}
        \centering
        \includegraphics[width=1.0\linewidth]{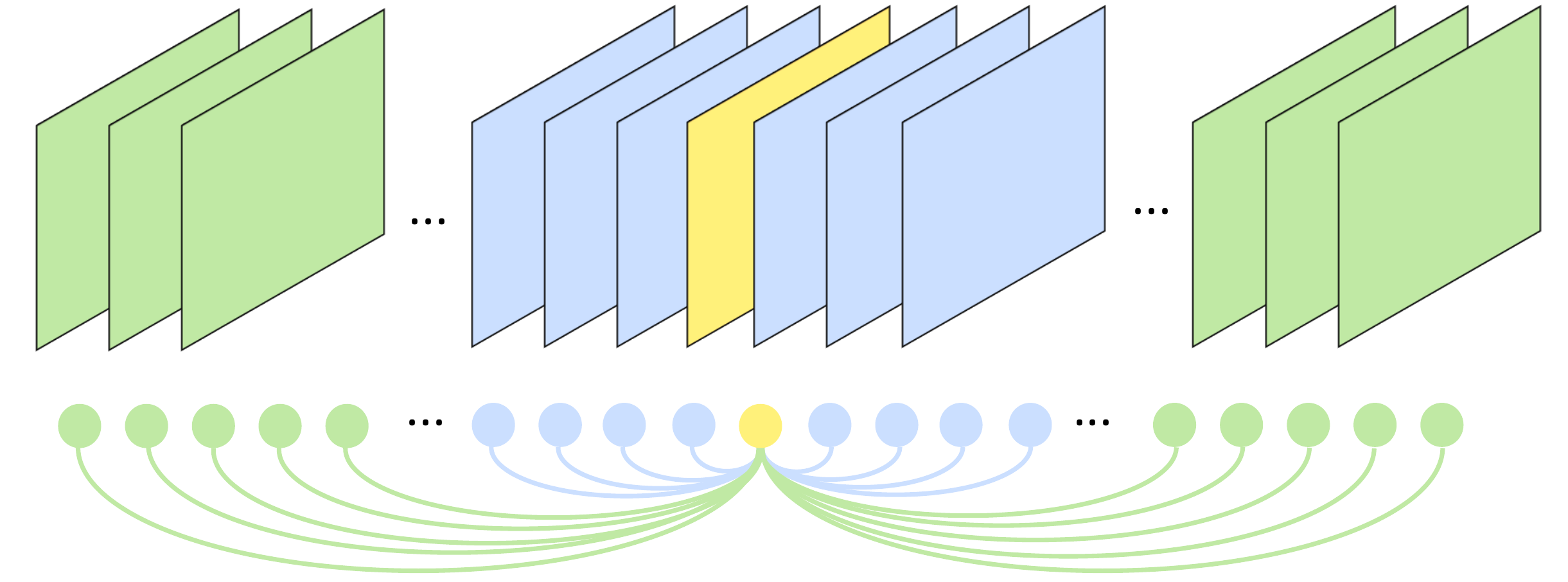}
    \end{minipage}
    \label{fig:fullview}
}
\subfigure[the aggregation size of local aggregation methods.]{
    \begin{minipage}[t]{0.9\linewidth}
        \centering
        \includegraphics[width=1.0\linewidth]{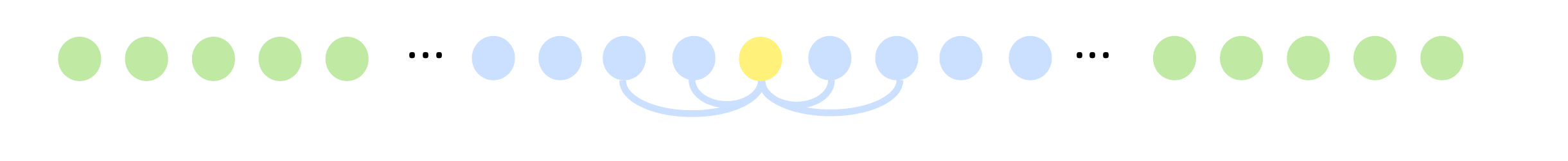}
    \end{minipage}
    \label{fig:localview}
}
\subfigure[the aggregation size of global aggregation methods.]{
    \begin{minipage}[t]{0.9\linewidth}
        \centering
        \includegraphics[width=1.0\linewidth]{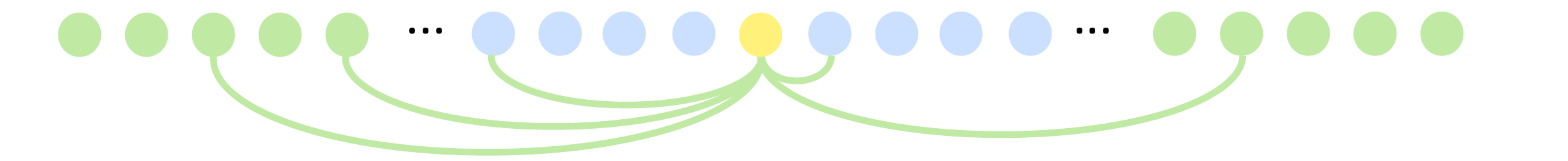}
    \end{minipage}
    \label{fig:globalview}
}
\subfigure[the aggregation size of our proposed methods.]{
    \begin{minipage}[t]{0.9\linewidth}
        \centering
        \includegraphics[width=1.0\linewidth]{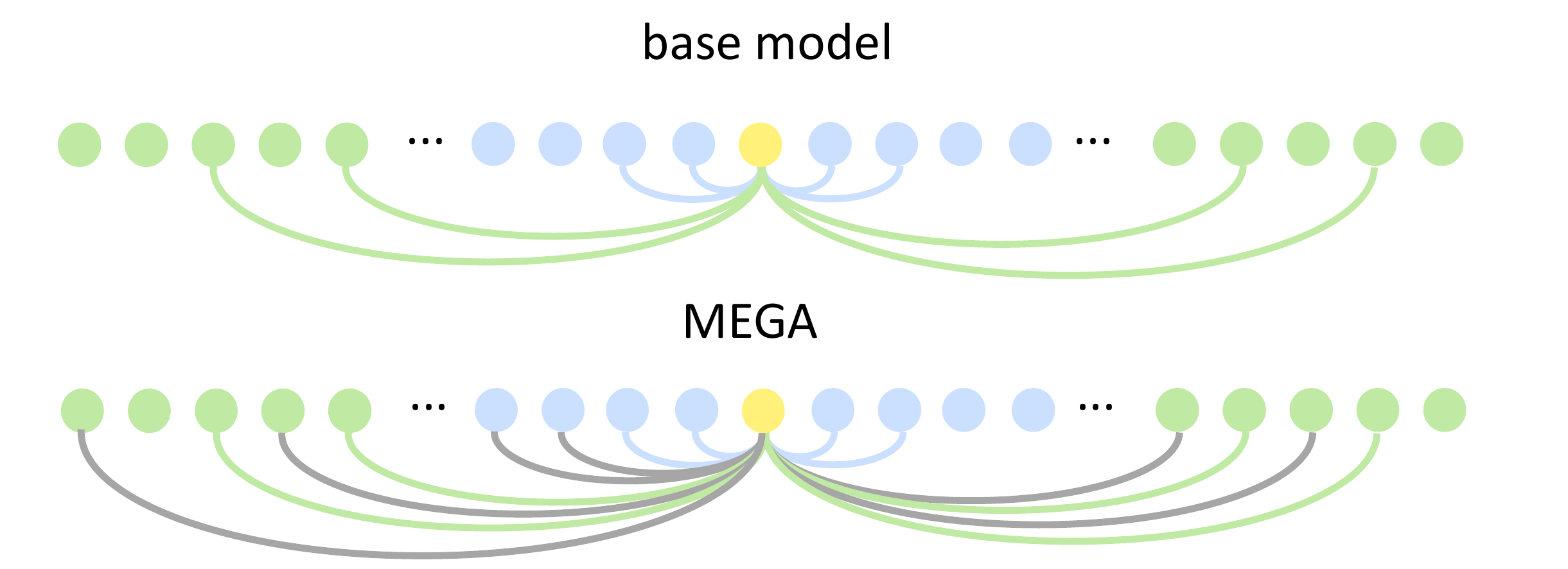}
    \end{minipage}
    \label{fig:ourview}
}
\subfigure{
    \begin{minipage}[t]{0.9\linewidth}
        \centering
        \includegraphics[width=1.0\linewidth]{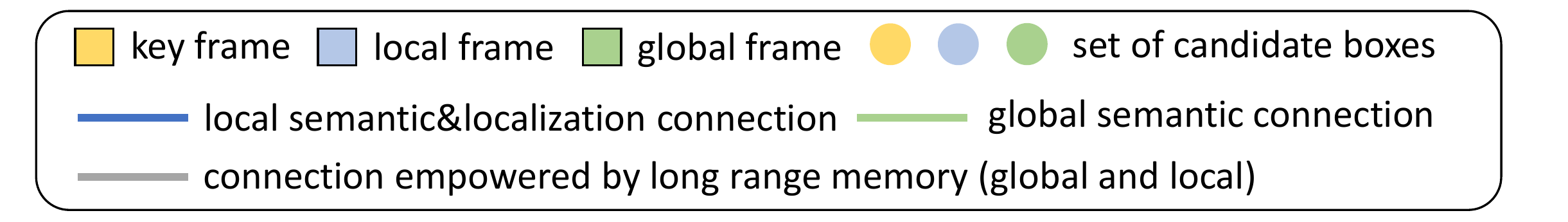}
    \end{minipage}
    \label{fig:legend}
}
\end{center}
   \caption{The aggregation size of different methods for addressing video object detection.}
\label{fig:graphview}
\end{figure}

What differs detecting objects in videos from detecting them in static images? A quick answer is the information lies in the temporal dimension. When isolated frame may suffer from problems such as motion blur, occlusion or out of focus, it is natural for humans to find clues from the entire video to identify objects.

When people are not certain about the identity of an object, they would seek to find a distinct object from other frames that shares high semantic similarity with current object and assign them together. We refer this clue as \textbf{global} semantic information for every frame in the video could be reference to. But semantic information alone fails in the case if we are not sure about whether an object exists or not, \eg, a black cat walking in the dark. We could not rely on the semantic information to tell us where it is, as the existence of the instance has not been approved in key frame yet. This problem could be alleviated if nearby frames were given. With information such as motion calculated by the difference between nearby frames, we could localize the object in key frame. We refer this source of information as \textbf{local} localization information. To be general, people identify objects mainly with these two sources of information.

Consequently, it is direct to borrow this idea from human to enhance the video object detection methods with the information inside the whole video, just as shown in Figure \ref{fig:fullview}. However, due to the huge amount of boxes exist in the whole video, enhancement with the information in the entire video is infeasible. This motivates us to perform approximation while keeping a balance between efficiency and accuracy. Recent methods towards solving video object detection could be viewed as different approaches of approximation and can be classified into two main categories: local aggregation methods and global aggregation methods. Methods like \cite{zhu17fgfa, feichtenhofer17dt, wang18manet, bertasius18stsn, deng19rdn} consider utilizing both semantic and localization information in a short local range just as shown in Figure \ref{fig:localview}. On the other hand, \cite{wu19selsa, deng2019ogemn, shvets19lltr} just consider the semantic impact between boxes, as shown in Figure \ref{fig:globalview}. Unfortunately, none of these methods takes a joint view of both local and global information. We refer this as the \textit{ineffective} problem.



Another problem existing in recent works is the size of frames for aggregation, which means the amount of information the key frame could gather from. In previous state-of-the-art methods \cite{zhu17fgfa, feichtenhofer17dt, wang18manet, bertasius18stsn, deng19rdn, wu19selsa}, only 20-30 reference frames are selected, which lasts only 1-2 seconds, for feature aggregation. This is also illustrated in Figure \ref{fig:localview} and \ref{fig:globalview}. We argue that the size of aggregation at this scale is an \textit{insufficient} approximation of either local influence or global influence, to say nothing of Figure \ref{fig:fullview}.


In this paper, we present Memory Enhanced Global-Local Aggregation (MEGA) to overcome the aforementioned \textit{ineffective} and \textit{insufficient} problem. Specifically, MEGA augments candidate box features of key frame by effectively aggregating global and local information. 

MEGA is instantiated as a multi-stage structure. In the first stage, MEGA aims to solve the \textit{ineffective} problem by aggregating both global and local information to key frames. However, just as shown in the top half of Figure \ref{fig:ourview}, the available content is still quite limited. So in the second stage, we introduce a novel \textit{Long Range Memory (LRM)} module which enables key frame to get access to much more content than any previous methods. In particular, instead of computing features from scratch for current key frame, we reuse the precomputed features obtained in the detection process of previous frames. These precomputed features are cached in \textit{LRM} and builds up a recurrent connection between current frame and previous frames. Unlike traditional memory, please note that these cached features are first enhanced by global information, which means current key frame is able to access more information not only locally, but also globally. The aggregation size is shown in the bottom half of Figure \ref{fig:ourview}. By introducing \textit{LRM}, a great leap towards solving the \textit{insufficient} problem is achieved while remaining simple and fast. 

Thanks to the enormous aggregation size of MEGA empowered by \textit{LRM}, we achieve 85.4\% mAP on ImageNet VID dataset, which is to-date the best reported result.

\section{Related Work}

\textbf{Object Detection from Images.} Current leading object detectors are build upon deep Convolutional Neural Networks (CNNs) \cite{krizhevsky12alexnet, simonyan14vgg, szegedy15inception, he16res, cao2019gcnet} and can be classified into two main familys, namely, anchor-based detectors (\eg, R-CNN \cite{rbg14rcnn}, Fast(er) R-CNN \cite{rbg15fastrcnn, ren15faster}, Cascade R-CNN \cite{cai18cascadercnn}) and anchor-free detectors (\eg, CornerNet \cite{law18cornernet}, ExtremeNet \cite{zhou19extreme}). Our method is built upon Faster-RCNN with ResNet-101, which is one of the state-of-the-art object detector.

\textbf{Object Detection in Videos.} Due to the complex manner of video variation, \eg, motion blur, occlusion and out of focus, it is not trivial to generalize the success of image detector into the video domain. The main focus of recent methods \cite{kang16TCNN, han16seqnms, zhu17dff, zhu17fgfa, zhu18hp, feichtenhofer17dt, wang18manet, bertasius18stsn, xiao18stmn, deng19rdn, wu19selsa} towards solving video object detection is improving the performance of per-frame detection by exploiting information in the temporal dimension. These methods could be categorized into local aggregation methods and global aggregation methods.


Local aggregation methods \cite{zhu17dff, zhu17fgfa, zhu18hp, feichtenhofer17dt, wang18manet, bertasius18stsn, xiao18stmn, deng19rdn} primarily utilize information in local temporal range to aid the detection of current frame. For example, FGFA \cite{zhu17fgfa}, MANet \cite{wang18manet} utilize optical flow predicted by FlowNet \cite{dosovitskiy15flownet} to propagate features across frames. Methods like STSN \cite{bertasius18stsn}, STMN \cite{xiao18stmn} directly learn to align and aggregate features without optical flow. Besides of these pixel-level aggregation methods, RDN \cite{deng19rdn} based on Relation Network \cite{hu18relationnet} directly learns the relation among candidate boxes of different frames in a local range to enhance the box-level features.

Global aggregation methods \cite{wu19selsa, deng2019ogemn, shvets19lltr} seek to enhance the pixel or box features directly with semantic information. Unlike optical flow or position relationship between boxes that depend on locality in temporal range to some extent, semantic similarity is somewhat independent of temporal distance. However, on the one hand getting rid of locality could enable model utilizing rich information beyond a fixed temporal window, on the other hand the lack of locality information would introduce weakness when localizing. 

Unlike these methods that separately view video object detection globally or locally, MEGA seeks to take full merit of both local and global aggregation to enhance the feature representation. Moreover, enabled by memory, information from longer content could be utilized.

\textbf{Information Aggregation beyond Local Range.} Like our method that tries to aggregate information beyond a small local range, \cite{wu19ltfb, oh19vosstmn, woo18linknet,xu2019spatial} share a similar spirit and shows superior results in different areas. \cite{wu19ltfb} also tries to aggregate information both globally and locally, however their "global" is just a larger local range while our "global" is the whole video. \cite{oh19vosstmn} keeps a carefully designed memory to aid video segmentation while ours memory is simpler and already efficient. \cite{woo18linknet} shares a similar relation aggregation module with ours. However how this module is instantiated is not the main focus of our work.

\section{Method}

\begin{figure*}[t]
\begin{center}
\subfigure[\textit{base model}]{
    \begin{minipage}[t]{0.38\linewidth}
        \centering
        \includegraphics[width=0.92\linewidth]{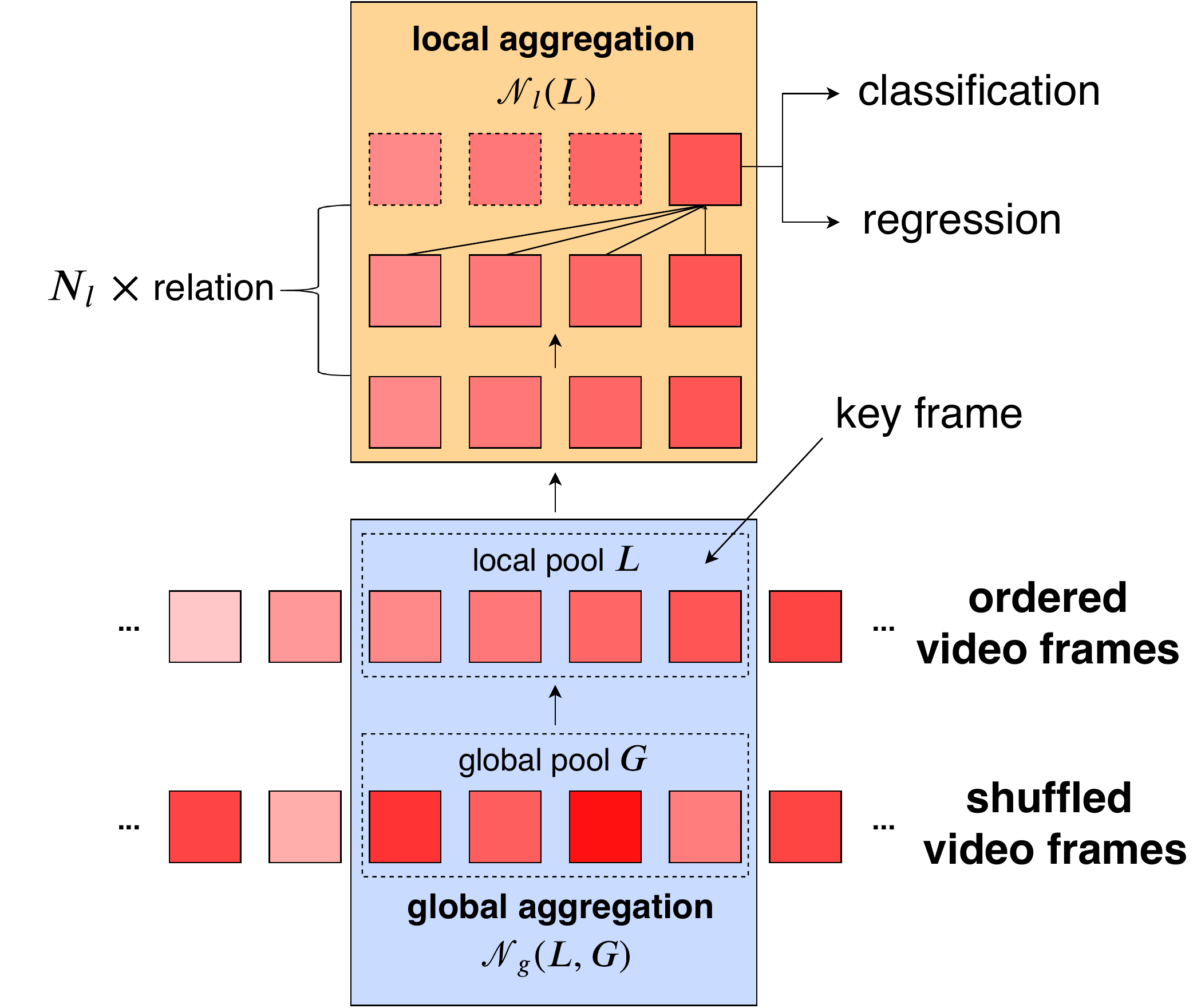}
    \end{minipage}
    \label{fig:methodbasic}
}\hspace{20pt}
\subfigure[MEGA]{
    \begin{minipage}[t]{0.48\linewidth}
        \centering
        \includegraphics[width=0.92\linewidth]{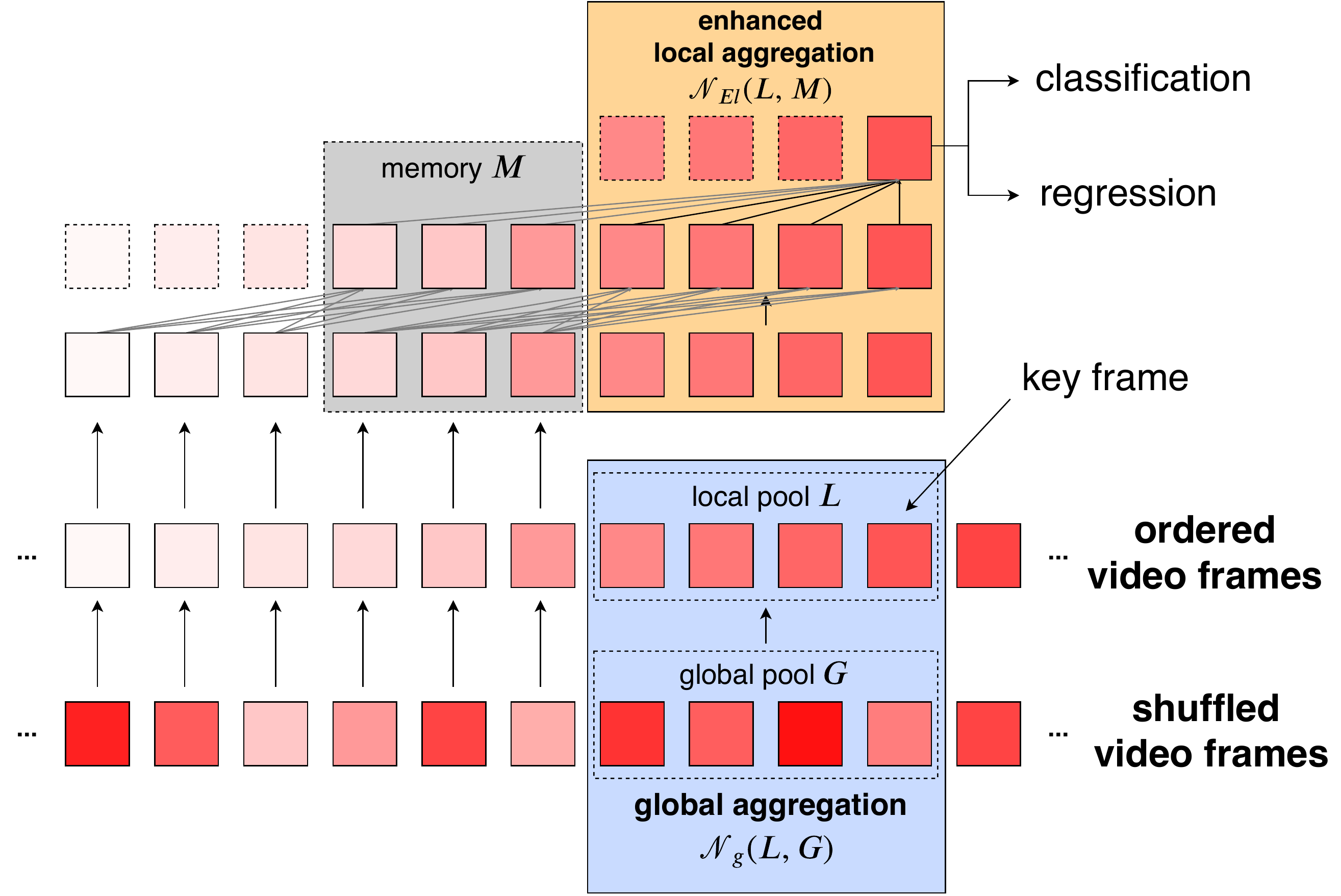}
    \end{minipage}
    \label{fig:methodfull}
}
\end{center}
    \caption{An overview of \textit{base model} and our proposed MEGA. For the purpose of illustration convenience, the key frame is placed on the very right. \textbf{(a) base model} with local pool size $T_l=4$, global pool size $T_g=4$, local aggregation stage $N_l=2$: we first aggregate global information from $\mathbf{G}$ into $\mathbf{L}$ (in blue box). In practice, this stage is instantiated as a stack of $N_g$ location-free relation modules. After that, $N_l$ location-based relation modules are utilized to mine the complex spatial-temporal information underlies $\mathbf{L}$ (in orange box). As can be seen, one frame is only capable to gather information from $T_l$ local reference frames and $T_g$ global reference frames. \textbf{(b) our proposed MEGA} with memory size $T_m=3$: the introduced novel long range memory module $\mathbf{M}$ (in gray box) enables the key frame could get information from far more frames than \textit{base model}. $\mathbf{M}$ caches precomputed intermediate features of local aggregation stack of previous frames. By utilizing these cached information and the recurrent connection empowered by $\mathbf{M}$, one frame is able to access information from totally $N_l\times T_m + T_l=10$ local reference frames and $N_l\times T_m + T_g=10$ global reference frames at this time, which is a great improvement over $T_l=4$ and $T_g=4$ of \textit{base model}. Additionally, as the cached information do not need any update, this makes the introduced computation overhead low. After the final enhanced features of current frame are produced, they will be propagated into traditional RCNN head to give classification and regression result.}
    \label{fig:method}
\end{figure*}

In this section, we will elaborate how we devise MEGA to enable the whole architecture to fully utilize both global and local information. To be specific, MEGA first aggregates selected global features to local features and afterwards, these global-enhanced local features together with a novel \textit{Long Range Memory (LRM)} module aggregate longer content of global and local information into key frame for better detection. An overview is depicted in Figure \ref{fig:methodfull}. 

\subsection{Preliminary}

The goal of video object detection is to give detection results for every frame of video $\{I_t\}_{t=1}^{T}$. Suppose the current frame to be detected is $I_k$ and $B_t=\{b_t^i\}$ denotes the candidate boxes generated by RPN at each frame $I_t$. All candidate boxes in adjacent frames $\{I_t\}_{t=k-\tau}^{k+\tau}$ are grouped together to form the local pool, that is, $\mathbf{L}=\left\{B_t\right\}_{t=k-\tau}^{k+\tau}$. For global features, we randomly shuffle the ordered index sequence $\{1,\ldots,T\}$ to obtain a shuffled index sequence $S$, then we sequentially select $T_g$ frames and group all boxes in them to form the global pool. It could be denoted as $\mathbf{G}=\left\{B_{S_i}\right\}_{i=k}^{k+T_g-1}$. At last, a novel long range memory module $\mathbf{M}$ stores intermediate features produced in the detection process of previous frame is introduced to allow the key frame
to exploit cached information, leading to an ability of modeling longer-term global and local dependency. Our ultimate goal is to give classification and regression results to all candidate boxes $B_k$ in the key frame with the help of $\mathbf{L},\mathbf{G}$ and $\mathbf{M}$.

Furthermore, we represent each box $b^i$ with its semantic feature $f_i$ and localization feature $g_i$. $g_i$ represents both the spatial information (\ie, height, width, center location) and the temporal information (\ie, the frame number).

\textbf{Relation Module.} The operator that we choose to mine the relationship between boxes is the relation module introduced in \cite{hu18relationnet} which is inspired by the multi-head attention \cite{Vaswani17attention}. Given a set of boxes $B=\{b_i\}$, object relation module is devised to enhance
each box $b_i$ by computing $M$ relation features as a weighted sum of semantic features from other boxes, where $M$ denotes the number of heads. Technically, the $m$-th relation features of $b_i$ is computed as
\begin{equation}
\small{f_{R}^{m,*}\left(b_{i}, B\right)=\sum_{j} \omega_{i j}^{m,*} \cdot\left(W_{V}^{m} \cdot f_{j}\right),~~ m=1, \cdots, M},
\end{equation}
where $W_V^m$ is a linear transformation matrix. The relation weight $\omega_{i j}^{m, *}$ indicates the impact between $b_i$ and $b_j$ measured with semantic features $f$ and probably localization features $g$. Here $* \in \{L, N\}$ denotes whether the localization features $g$ is incorporated in $\omega$, where $L$ means incorporated and $N$ means not. As localization features between two distant boxes in the temporal dimension are redundant and may harm the overall performance, we design the location-free version urging the relation module to focus only on semantic features. Notice that unlike \cite{hu18relationnet, deng19rdn}, temporal information is also contained in our localization features to distinguish the impact of boxes from different frames. This temporal information is incorporated in $\omega$ in a relative manner as in \cite{dai19transxl}.

Finally, by concatenating all $M$ relation features and its original feature, we obtain the output augmented feature:
\begin{equation}
f_{r m}^*\left(b_{i}, B\right)=f_{i}+\operatorname{concat}\left[\left\{f_{R}^{m,*}\left(b_{i}, B\right)\right\}_{m=1}^{M}\right], \label{eq:relation}
\end{equation}
where $* \in \{L, N\}$ and the meaning is the same as before.

After the augmented feature is produced, we additionally append a non-linear transformation function $h(\cdot)$ implemented as a fully-connected layer and ReLU after it.

Further, We could extend the relation module to modeling the relationship between two sets of boxes. For convenience, we use notation $f_{rm}^*(B,P)$ to represent the collection of all augmented proposal features, \ie, $\{f_{r m}^*\left(b_{i}, P\right)\}$, which means all the bounding boxes in $\mathbf{B}$ are augmented via the features of bounding boxes in $\mathbf{P}$.

\subsection{Memory Enhanced Global-Local Aggregation} \label{sec:MEGA}

\textbf{Global-Local Aggregation for \textit{ineffective} problem.} First we will elaborate how we design the network by aggregating global and local features together to address the \textit{ineffective} problem, which means separately considering global or local information. We denote this architecture as \textit{base model} and it is depicted in Figure \ref{fig:methodbasic}. 

Specifically, the global features from $\mathbf{G}$ are aggregated into $\mathbf{L}$ at first. The update function could be formulated as:
\begin{gather}
\mathbf{L}^g=\mathcal{N}_g(\mathbf{L},\mathbf{G}), \label{eq:globalstage}
\end{gather}
where $\mathcal{N}_g(\cdot)$ is a function composed of stacked location-free relation modules and $\mathbf{L}^g$ denotes the final global-enhanced version of $\mathbf{L}$ after this function. As our aim is to fully exploit the potential of global features to enhance local features, we iterate the relation reasoning in a stacked
manner equipped with $N_g$ relation modules to better characterize the relations between $\mathbf{G}$ and $\mathbf{L}$. To be specific, the calculation process in the $k$-th relation module is given as:
\begin{gather}
\mathbf{L}^{g,k}=f_{rm}^{N}(\mathbf{L}^{g,k-1}, \mathbf{G}),~~~~~ k=1,...,N_g,
\end{gather}
where $f_{rm}^{N}(\cdot)$ denotes location-free relation module, defined in Eq \eqref{eq:relation} and $\mathbf{L}^{g,0}=\mathbf{L}$ denotes the input of the first relation module. Latter relation module takes the output from previous relation module as input. Finally, the output of the $N_g$-th relation module is taken as $\mathbf{L}^g$.

After the global features are aggregated into local features, we then seek to utilize semantic and localization information underlies these local features to further enhance themselves. To achieve this, a stack of $N_l$ location-based relation modules is adopted. Technically, the overall function could be summarized as:
\begin{equation}
{\mathbf{L}}^l=\mathcal{N}_l(\mathbf{L}^g), \label{eq:localstage}
\end{equation}
where $\mathbf{L}^l$ represents the final enhanced version of local pool. We decompose the whole procedure of $\mathcal{N}_l(\cdot)$ in below. The computation pipeline in the $k$-th relation module is similar with its counterpart in $\mathcal{N}_g(\cdot)$:
\begin{gather}
\mathbf{L}^{l,k}=f_{rm}^{L}(\mathbf{L}^{l,k-1}, \mathbf{L}^{l,k-1}), ~~~~~k=1,...,N_l,
\end{gather}
where $f_{rm}^{L}(\cdot)$ denotes location-based relation module and we adopt $\mathbf{L}^g$, the global-enhanced version of $\mathbf{L}$, as the input of the first location-based relation module. $\mathbf{L}^{l,N_l}$ is taken as the output enhanced pool $\mathbf{L}^l$. After the final update is done, all box features in $\mathbf{L}^l$ that belong to the key frame will be extracted and be propagated through traditional RCNN head to give classification and regression result. These extracted features is denoted as $\mathbf{C}$.

\textbf{Long Range Memory for \textit{insufficient} problem.} With \textit{base model}, a single frame is able to aggregate totally $T_g$ frames of global features and $T_l$ frames of local features as shown in Figure \ref{fig:methodbasic}, which is a big step towards solving the \textit{ineffective} problem. But the \textit{insufficient} problem, which stands for the size of frames for key frame to aggregate is too small, is still leaving untouched until now. This problem can be naively solved by increasing $T_g$ and $T_l$ to approach the length of the video if infinite memory and computation is available. However, this is infeasible because the resource is limited in practice.

So how could we solve the \textit{insufficient} problem while keeping the computation cost affordable? Inspired by the recurrence mechanism introduced in \cite{dai19transxl}, we design a novel module named \textit{Long Range Memory (LRM)} to fulfill this target. To summarize, \textit{LRM} empowers \textit{base model} to capture content from much longer content, both globally and locally, by taking full use of precomputed features. We name this memory enhanced version as MEGA. We refer readers to Figure \ref{fig:methodfull} for an overview of how MEGA works.

To see the defect of \textit{base model}, suppose $I_{k-1}$ and $I_k$ are two consecutive frames. 
When we shift the detection process from $I_{k-1}$ to $I_{k}$, we discard all intermediate features produced by the detection on $I_{k-1}$. So the detection process of $I_k$ could not take any merit of the detection process of $I_{k-1}$, though they are consecutive in the temporal dimension. Every time we move to a new frame, we need to recompute everything from the very start. This motivates us to memorize precomputed features to allow current frame to exploit more information in the history. In practice, besides utilizing information from adjacent frames $\{I_t\}_{t=k-\tau}^{k+\tau}$, long range memory $\mathbf{M}$ of size $T_m$ would additionally provide features of $T_m$ frames of information ahead of adjacent frames, namely $\{I_t\}_{t=k-\tau-T_m}^{k-\tau-1}$, to help detecting on $I_k$.

To be specific, after detection process for $I_{k-1}$ is done, Unlike in \textit{base model} where we discard all features computed in the detection process, at this time intermediate features of $I_{k-\tau-1}$, the first frame of adjacent frames of $I_{k-1}$, would be cached in long range memory $\mathbf{M}$. That means, as our local aggregation function $\mathcal{N}_l(\cdot)$ defined in Eq \eqref{eq:localstage} is composed of a stack of $N_l$ relation modules and $\mathbf{L}^{l, i}, i \in \{0, N_l\}$ is the features enhanced after the $i$-th relation module ($i=0$ denotes the input), we would extract and store all level of features of $I_{k-\tau-1}$, namely $\mathbf{L}_{k-\tau-1}^{l, i},~i \in \{0, N_l\}$ in $\mathbf{M}$. Concretely, $\mathbf{M}$ is of $N_l+1$ levels where $\mathbf{M}^i$ caches $\mathbf{L}_{k-\tau-1}^{l, i}$. Every time the detection process is done for a new frame, features corresponding to the first frame of adjacent frames of this new frame would be added into $\mathbf{M}$. 

\begin{algorithm}[t]
\footnotesize
\caption{Inference Algorithm of MEGA}
\label{Alg:MEGA}
\begin{algorithmic}[1]
\STATE \textbf{Input:} video frames $\{I_t\}$ of length $T$
\STATE \textbf{Initialize:} long range memory $\mathbf{M}$ to be empty
\FOR {$t = 1$ to $\tau+1$}
    \STATE $B_t=\mathcal{N}_{RPN}(I_t)$ ~~~~~~~~~~~~~~~~~~~~~// generate candidate boxes of $I_t$
\ENDFOR
\STATE // shuffle the frame index for random global feature selection
\STATE $S$ = $\operatorname{\textbf{shuffle}}(1 \ldots T)$
\FOR {$t = 1$ to $T_g$}
    \STATE $B_{S_t}=\mathcal{N}_{RPN}(I_{S_t})$ ~~~~~~~~~~~~~~~// generate candidate boxes of $I_{S_t}$
\ENDFOR
\FOR {$k = 1$ to $T$}
    \STATE $\mathbf{L}=\{B_t\}_{t=k-\tau}^{ k+\tau}$ ~~~~~~~~~~~~~~// form local pool with adjacent frames
    \STATE $\mathbf{G}=\{B_{S_t}\}_{t=k}^{k+T_g-1}$ ~~~~~~~// form global pool with random frames
    \STATE $\mathbf{L}^g=\mathcal{N}_g(\mathbf{L},\mathbf{G})$~~~~~~~~~~~~~~~~~~~~~~~~~~~ // global agg. stage with Eq \eqref{eq:globalstage}
    \STATE $\mathbf{L}^l=\mathcal{N}_{El}(\mathbf{L}^g, \mathbf{M})$ ~~~~~~~~// enhanced local agg. stage with Eq \eqref{eq:elocalstage}
    \STATE $\mathbf{C}=$\text{Select}-$I_k(\mathbf{L}^l)$ ~~~~// extract enhanced features of key frame $I_k$
    \STATE $\mathbf{D}_k=\mathcal{N}_{RCNN}(\mathbf{C})$ ~~~~~~~~~~~~~~~~~~~~~~~~~~~~~~~// detect on the key frame
    \STATE \textbf{Update}$(\mathbf{M}, \mathbf{L}_{k-\tau}^{l,*})$ ~~~~~~~~~~~~~~~~~~~~~~// update the long range memory
    \STATE $\mathbf{B}_{k+\tau+1}=\mathcal{N}_{RPN}(I_{k+\tau+1})$
    \STATE $\mathbf{B}_{S_{k+T_g}}=\mathcal{N}_{RPN}(I_{S_{k+T_g}})$
\ENDFOR
\STATE \textbf{Output:} The final detection result of the video: $\{\mathbf{D}_k\}$
\end{algorithmic}
\end{algorithm}

By introducing $\mathbf{M}$ and reusing these precomputed features cached in it, we could enhance our local aggregation stage to incorporate the information from $\mathbf{M}$ for detection at $I_k$ and latter frames. The enhanced version of local aggregation could be summarized as
\begin{equation}
{\mathbf{L}}^l=\mathcal{N}_{El}(\mathbf{L}^g, \mathbf{M}) \label{eq:elocalstage}
\end{equation}

Like $\mathcal{N}_{l}(\cdot)$, $\mathcal{N}_{El}(\cdot)$ is also build upon a stack of $N_l$ location-based relation module while taking the impact of $\mathbf{M}$ into consideration:
\begin{gather}
\mathbf{L}^{l,k}=f_{rm}^{L}(\mathbf{L}^{l,k-1}, [\mathbf{L}^{l,k-1},\mathbf{M}^{k-1}]), k=1,...,N_l,
\end{gather}
where $[\cdot,\cdot]$ indicates the concatenation of two information pools. Compared to the standard update function, the critical difference lies in the formation of the reference pool. Like in \textit{base model}, after the final update is done, $\mathbf{C}$ will be extracted and be propagated through traditional RCNN head to give classification and regression result of current key frame. The detailed inference process of MEGA is given in Algorithm \ref{Alg:MEGA}. 


\textbf{To what extent does \textit{LRM} address the \textit{ineffective} and \textit{insufficient} approximation problem?} By appending long range memory $\mathbf{M}$ of size $T_m$, a direct increase of $T_m$ in feature number is obvious. But the increase of sight is far beyond this number. Please note that our enhanced local stage is of $N_l$ stacks, thanks to the recurrent connection introduced by $\mathbf{M}$, the key frame could gather information from additional $T_m$ frames every time we iterate the relation reasoning just as shown in Figure \ref{fig:methodfull}. The last and most importantly, as the cached features of every frame are first enhanced by different set of global features, long range memory does not only increase the aggregation size locally, but also globally. To summarize, model with $N_l$-level enhanced local aggregation stage could gather information from totally $N_l \times T_m + T_l$ local reference frames and $N_l \times T_m + T_g$ global reference frames, where $T_l, T_g, T_m$ denotes the size of local pool, global pool and memory, respectively. It is a great leap towards $T_l$ and $T_g$ in the \textit{base model}. With this hugely enlarged aggregation size, we argue the \textit{ineffective} and \textit{insufficient} problem is better addressed by our model while not increasing the running time too much. This is further justified by the superior experimental results shown in Table \ref{table:mainresult1} in Section \ref{sec:experiment}. 


\section{Experiments} \label{sec:experiment}

\subsection{Dataset and Evaluation Setup}

We evaluate our proposed methods on ImageNet VID dataset \cite{RussakovskyDSKS15}. The ImageNet VID dataset is a large scale benchmark for video object detection task consisting of 3,862 videos in the training set and 555 videos in the validation set. 30 object categories are contained in this dataset. Following protocols widely adopted in \cite{zhu17fgfa, zhu17dff, wang18manet, zhu18hp}, we evaluate our method on the validation set and use mean average precision (mAP) as the evaluation metric.

\subsection{Network Architecture}

\textbf{Feature Extractor.} We mainly use ResNet-101\cite{he16res} and ResNeXt-101 \cite{xie17resnext} as the feature extractor. As a common practice in \cite{zhu17fgfa, wang18manet, deng19rdn, wu19selsa}, we enlarge the resolution of feature maps by modifying the stride of the first convolution block in last stage of convolution, namely \textit{conv5}, from 2 to 1. To retain the receptive field size, the dilation of these convolutional layers is set as 2.

\textbf{Detection Network.} We use Faster R-CNN \cite{ren15faster} as our detection module. The RPN head is added on the top of \textit{conv4} stage. In RPN, the anchors are of 3 aspect ratios \{1:2, 1:1, 2:1\} and 4 scales \{$64^2, 128^2, 256^2, 512^2$\}, resulting in 12 anchors for each spatial location. During training and inference, $N=300$ candidate boxes are generated for each frame at an NMS threshold of 0.7 IoU. After boxes are generated, we apply RoI-Align \cite{he17maskrcnn} and a 1024-D fully-connected layer after \textit{conv5} stage to extract RoI feature for each box.

\textbf{MEGA.} At training and inference stages, the local temporal window size is set to $T_l=25$ ($\tau=12$). Note that the temporal span could be different on two sides of key frame in practice. For the sake of efficiency, we do not keep all candidate boxes produced by RPN for each local reference frame, instead $80$ candidate boxes with highest objectness scores are selected. And in the latter stack of local aggregation stage, the number of boxes are further reduced to $20$. As for the global reference frame, we totally select $T_g=10$ frames and $80$ proposals with highest objectness scores each frame. The size $T_m$ of Long Range Memory is set to $25$.

As for the global and local aggregation stage, the number of relation modules is set to $N_g=1$ and $N_l=3$, respectively. For each relation module, the hyperparameter setting is the same as \cite{hu18relationnet}.

\subsection{Implementation Details}
Following common protocols in \cite{zhu17fgfa, wang18manet, deng19rdn, wu19selsa}, we train our model on a combination of ImageNet VID and DET datasets. For the DET dataset, we select images that are of the same 30 classes as in the VID dataset. We implement MEGA mainly on maskrcnn-benchmark \cite{massa2018mrcnn}. The iuput images are resized to have their shorter side to be 600 pixels. The whole architecture is trained on 4 RTX 2080ti GPUs with SGD. Each GPU holds one mini-batch and each mini-batch contains one set of images or frames. We train our network for totally 120K iteration, with learning rate $10^{-3}$ and $10^{-4}$ in the first 80K and in the last 40K iterations, respectively. At inference, an NMS of 0.5 IoU threshold is adopted to suppress reduplicate detection boxes.

\textbf{Training.} As the number of boxes prohibits the training process to be the same as the inference process, we adopt the strategy of temporal dropout\cite{zhu17fgfa} to train our model. Given key frame $I_k$, we randomly sample two frames from $\{I_t\}_{t=k-\tau}^{k+\tau}$ and two frames from the whole video to approximately form $\mathbf{L},\mathbf{G}$. Additional two frames from $\{I_t\}_{t=k-\tau-T_m}^{k-\tau-1}$ are selected to construct $\mathbf{M}$. For convenience, we name them as $\hat{\mathbf{L}},\hat{\mathbf{G}}$ and $\hat{\mathbf{L}}_M$. We first apply \textit{base model} on $\hat{\mathbf{G}}$ and $\hat{\mathbf{L}}_M$ and store all intermediate features produced as $\hat{\mathbf{M}}$. After that, $\hat{\mathbf{L}},\hat{\mathbf{G}},\hat{\mathbf{M}}$ are propagated through full MEGA to generate $\mathbf{C}$. Finally, the whole model are trained with classification and regression losses
over $\mathbf{C}$. One thing needed to be pointed out is that we stop the gradient flow in the construction of $\hat{\mathbf{M}}$. This behavior is similar to \cite{dai19transxl} but with different motivation: we would like the model to pay more attention on most adjacent frames.

\subsection{Main Results}

\begin{table}
\footnotesize
\begin{center}
    \addtolength{\tabcolsep}{-1.5pt}
\begin{tabular}{c|c|cc|c}
\Xhline{1.0pt}
Methods & Backbone & local & global & mAP(\%) \\
\hline
FGFA \cite{zhu17fgfa} & ResNet-101 & \checkmark& &76.3\\
MANet \cite{wang18manet} & ResNet-101 & \checkmark& &78.1\\
THP \cite{zhu18hp} & ResNet-101+DCN & \checkmark& &78.6\\
STSN \cite{bertasius18stsn} & ResNet-101+DCN & \checkmark& &78.9\\
OGEMN \cite{deng2019ogemn} & ResNet-101+DCN & & \checkmark & 80.0 \\
SELSA \cite{wu19selsa} & ResNet-101 & & \checkmark & 80.3\\
RDN \cite{deng19rdn} & ResNet-101 & \checkmark & & 81.8\\
RDN \cite{deng19rdn} & ResNeXt-101 &\checkmark&& 83.2\\
\hline
\multirow{2}{*}{MEGA (ours)} & ResNet-101 & \checkmark& \checkmark & \textbf{82.9}\\
~ & ResNeXt-101 &\checkmark&\checkmark& \textbf{84.1}\\
\Xhline{1.0pt}
\end{tabular}
\end{center}
\caption{Performance comparison with state-of-the-art end-to-end video object detection models on ImageNet VID validation set.}
\label{table:mainresult1}
\end{table}

\begin{table}
\footnotesize
\begin{center}
\begin{tabular}{c|c|c}
\Xhline{1.0pt}
Methods & Backbone & mAP(\%) \\
\hline
FGFA \cite{zhu17fgfa} & ResNet-101 & 78.4\\
ST-Lattice \cite{chen18stlattice} & ResNet-101 & 79.6\\
MANet \cite{wang18manet} & ResNet-101 & 80.3\\
D\&T \cite{feichtenhofer17dt} & ResNet-101 & 80.2\\
STSN \cite{bertasius18stsn} & ResNet-101+DCN & 80.4\\
STMN \cite{xiao18stmn} & ResNet-101 & 80.5\\
SELSA \cite{wu19selsa} & ResNet-101 & 80.5\\
OGEMN \cite{deng2019ogemn} & ResNet-101+DCN & 81.6 \\
RDN \cite{deng19rdn} & ResNet-101 & 83.8\\
\Xhline{0.8pt}
MEGA (ours) & ResNet-101 & \textbf{84.5}\\
\Xhline{0.8pt}
FGFA \cite{zhu17fgfa} & Inception-ResNet & 80.1\\
D\&T \cite{feichtenhofer17dt} & Inception-v4 & 82.0\\
RDN \cite{deng19rdn} & ResNeXt-101 & 84.7\\
\Xhline{0.8pt}
MEGA (ours) & ResNeXt-101 & \textbf{85.4}\\
\Xhline{1.0pt}
\end{tabular}
\end{center}
\caption{Performance comparison with state-of-the-art video object detection models with post-processing methods (\eg Seq-NMS, Tube Rescoring, BLR).}
\label{table:mainresult2}
\end{table}

Table \ref{table:mainresult1} shows the result comparison between state-of-the-art end-to-end models without any post-processing. Among all methods, MEGA achieves the best performance and is the only method that takes full use of both global and local information. With ResNet-101 backbone, MEGA can achieve 82.9\% mAP, 1.1\% absolute improvement over the strongest competitor RDN. By replacing the backbone feature extractor from ResNet-101 to a stronger backbone ResNeXt-101, our method achieves better performance of 84.1\% mAP just as expected. Among all competitors, RDN is the most representative method in local aggregation scheme while SELSA is the one in global aggregation scheme. RDN only models the relationship within a short local temporal range, while SELSA only builds sparse global connection. As discussed in Section \ref{sec:intro}, these methods may suffer from the \textit{ineffective} and \textit{insufficient} approximation problems which lead to inferior results than ours.

Like many previous methods that could get further improvement by post processing, it could also benefit ours. Here the post processing technique we adopt is BLR \cite{deng19rdn}, which is done by finding optimal paths in the whole video and then re-score detection boxes in each path. Table \ref{table:mainresult2} summarizes the results of state-of-the-art methods with different post-processing techniques. With no doubt, our method still performs the best, obtaining 84.5\% and 85.4\% mAP with backbone ResNet-101 and ResNeXt-101, respectively.

\subsection{Ablation Study}

To examine the impact of the key components in our MEGA, we conduct extensive experiments to study how they contribute to our final performance.

\textbf{Long Range Memory.} We would first explore the effect of \textit{LRM} as it plays the key role in our full model. We show the performance comparison results between \textit{base model} and MEGA in Table \ref{table:ablationLRM}. As shown in the table, a gap of 1.5\% mAP exists between these two models, which is a significant improvement. Gap of 1\% mAP still exists after increasing \textit{base model}'s local span $\tau$ to $\tau+\frac{T_m}{2}$, while running at a much slower speed. We argue the superior performance of MEGA is brought by the novel \textit{LRM} module which enables one frame could gather information efficiently from much longer content, both globally and locally.


\begin{table}
\scriptsize
\begin{center}
\begin{tabular}{c|ccc|c|c}
\Xhline{1.0pt}
Methods & local & global & memory & mAP(\%) & runtime(ms) \\
\hline
single frame & \checkmark & & &75.4 & 64\\
\textit{base model} & \checkmark & \checkmark& &81.4 & 105.6\\
+ larger local range & \checkmark & \checkmark& &81.9 & 130\\\hline
MEGA & \checkmark& \checkmark& \checkmark&\textbf{82.9} & 114.5\\
\Xhline{1.0pt}
\end{tabular}
\end{center}
\caption{Performance of \textit{base model} and MEGA.}
\label{table:ablationLRM}
\end{table}

Table \ref{table:ablationGlobalLocal} shows the results of removing global or local information from MEGA. In the default setting, the number of relation modules in the global aggregation stage and local aggregation stage is $N_g=1$ and $N_l=3$, respectively. To study the effect of global features, We simply set $N_g$ to $0$ to remove the influence from it. As shown in the table, MEGA experiences a sheer performance loss of 1.6\% mAP by removing the global features. We also conduct an experiment by setting $N_l$ to $4$ which means to keep the number of parameters as the same as MEGA though it is not fully comparable (the local range is larger). And this experiment gives 81.6\% mAP which is still lower. The above results show the importance of global feature aggregation. 

To see the importance of local information, we conduct an experiment by setting $N_g$ to 4 and $N_l$ to 0, also the local temporal window size $T_l$ and the number of global reference frames $T_g$ is changed to $1$ and $25$, respectively. Under this setting, one frame could only enhanced by global information while keeping the number of parameters the same as MEGA. The result is given in the last row of Table \ref{table:ablationGlobalLocal}. As can be seen, this result is 1.1\% lower than our full model, indicating the necessity of local features.


\begin{table}
\footnotesize
\begin{center}
\begin{tabular}{c|cc|c}
\Xhline{1.0pt}
Method & $N_g$ & $N_l$ & mAP(\%) \\
\hline
MEGA & 1 & 3 & \textbf{82.9}\\\hline
\multirow{2}{*}{MEGA (no global stage)} & 0 & 3 & 81.3\\
 & 0 & 4 & 81.6\\\hline
MEGA (no local stage) & 4 & 0 & 81.8\\
\Xhline{1.0pt}
\end{tabular}
\end{center}
\caption{Ablation study on the global and local feature aggregation. $N_l$ and $N_g$ is the number of relation modules in local aggregation stage and global aggregation stage, respectively. By setting $N_l$ or $N_g$ to 0 removes the influence of local or global information.}
\label{table:ablationGlobalLocal}
\end{table}

\begin{figure*}[ht]
\begin{center}
   \includegraphics[width=0.8\linewidth]{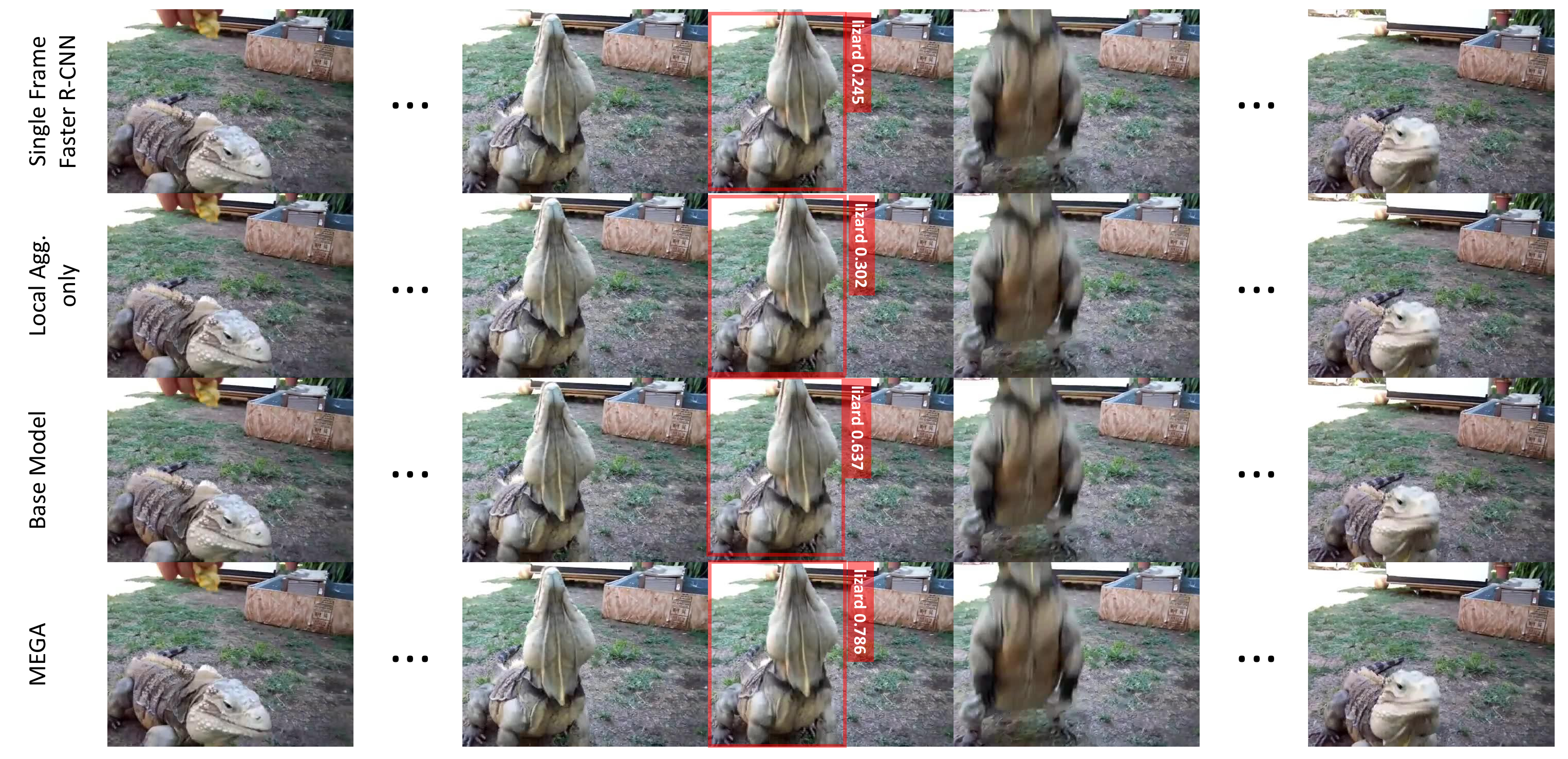}
\end{center}
   \caption{Example detection results of methods that incorporates different amount of local and global information.}
\label{fig:visualization}
\end{figure*}

From Table \ref{table:ablationLRM} and \ref{table:ablationGlobalLocal}, an interesting conclusion could be drawn. It is the combination of local features, global features and memory that empowers MEGA to obtain superior result, and three components alone is not sufficient. With memory, one frame could get access to more global and local features and in turn, this enhanced frame could offer us a more compact memory. This manner justifies our intuition that a better approximation of Figure \ref{fig:fullview} is a promising direction to boost the performance on video object detection.


Figure \ref{fig:visualization} showcases one hard example in video object detection. As the lizard is presented in a rare pose for quite a while, only exploiting local information is not able to tell us what it is. The top two rows shows the result of single frame and local aggregation. They are not able to recognize the object with just local information. By taking global features into consideration, the model overcomes this difficult case by aggregating features from distinct object from global frames. The last two rows show the result of \textit{base model} and MEGA. The result gets better when more information is incorporated just as expected.

\textbf{Aggregation Scale.} The aggregation scale here means the content one frame could gather from, globally or locally. Totally four hyperparameters, global reference frame number $T_g$, local reference frame number $T_l$, number of relation modules in local aggregation stage $N_l$ and memory size $T_m$ would influence the aggregation scale. The result is given in Table \ref{tab:ablation_num}. (a) The result of different number of $T_g$. As can be seen, there is only minor difference between them. As discussed in Section \ref{sec:MEGA}, the number of global reference frames one frame could see is $N_l \times T_m + T_g$, which indicates the influence from $T_g$ is minor compared to $N_l$ and $T_m$. (b) As for $T_l$, the result is similar with $T_g$. (c)(d) The result for $N_l$ and $T_m$. These two parameters jointly influence the global and local range of MEGA. The performance of MEGA gets worse when $N_l$ or $T_m$ is small, which implies larger aggregation scale matters. The improvement saturates when $N_l$ or $T_m$ gets bigger, which may suggest that a proper approximation is sufficient enough.

\begin{table}[ht]
    \footnotesize
    \begin{center}
    \addtolength{\tabcolsep}{1.5pt}
    \begin{tabular}{c|ccccc}
\Xhline{1.0pt}
\multicolumn{6}{c}{(a) {Global reference frame number $T_g$}}\\
\hline
$T_g$ &  5 & 10* & 15 & 20\\
mAP (\%) & 82.7 & 82.9 & 82.9 & 83.0\\
runtime (ms) & 111.6 & 114.5 & 117.4 & 124.2\\
\Xhline{1.0pt}
\multicolumn{6}{c}{(b) {Local reference frame number $T_l$}}\\
\hline
$T_l$ &  13 & 17 & 21 & 25* & 29\\
mAP (\%) & 82.6 & 82.7 & 82.8 & 82.9 & 82.9\\
runtime (ms) & 99.2 & 105.9 & 109.7 & 114.5 & 122.1\\
\Xhline{1.0pt}
\multicolumn{6}{c}{(c) {number of relation modules in local agg. stage $N_l$}}\\
\hline
$N_l$ &  1 & 2 & 3* & 4\\
mAP (\%) & 82.1 & 82.5 & 82.9 & 83.0\\
runtime (ms) & 100.6 & 108.7 & 114.5 & 122.3\\
\Xhline{1.0pt}
\multicolumn{6}{c}{(d) {Memory size $T_m$}}\\
\hline
$T_m$ &  5 & 15 & 25* & 35 & 45\\
mAP (\%) & 82.0 & 82.3 & 82.9 & 82.9 & 83.0\\
runtime (ms) & 111.3 & 113 & 114.5 & 115.4& 116.1\\
\Xhline{1.0pt}
    \end{tabular}
    \end{center}
    \caption{Ablation study on different global reference frame number $T_g$, local reference frame number $T_l$, number of relation modules in local aggregation stage $N_l$ and memory size $T_m$. Default parameter is indicated by *.}
    \label{tab:ablation_num}
\end{table}

\textbf{Types of Relation Modules.} As we has discussed earlier, we differentiate two types of relation modules by whether incorporating the location information into the relation weights. By incorporating location information in global aggregation stage, we obtain inferior result of 82.5\% mAP, which verifies that incorporating the location information to the global aggregation stage would harm the overall performance.

\textbf{Online Setting.} How would our model behave if current frame could only access information from previous frames? We refer this setting as online setting, which is an often confronted situation in practice. In experiment, we still keep the number of local reference frames $T_l$ to be $25$ while modifying the adjacent frames of key frame $I_k$ to be $\{I_t\}_{t=k-T_l+1}^k$. As for global information, the sampling range is limited to all previous frames of current frame while keeping global sampling number $T_g$ the same as before. With this limited sight, the modified MEGA could also obtain 81.9\% mAP. To the best of our knowledge, the previous state-of-the-art performance under the same setting is OGEMN's 80.0\% mAP, which uses a variant of memory network to boost detection. As memory network could be treated as a way to build sparse connection among different frames, it could be classified into global aggregation scheme. This method also suffers from the \textit{ineffective} and \textit{insufficient} problem, which results in inferior performance than ours.

\textbf{Running Time.} The running time of our model is given in Table \ref{table:ablationLRM}. Compared to the \textit{base model}, MEGA boosts the overall performance by a large margin while only introducing little computation overhead. This result suggests the efficiency of our method.


\section{Conclusion}

In this work, we present Memory Enhance Global-Local Aggregation Network (MEGA), which takes a joint view of both global and local information aggregation to solve video object detection. Particularly, we first thoroughly analyze the \textit{ineffecitve} and \textit{insufficient} problems existed in recent methods. Then we propose to solve these problems in a two-stage manner, in the first stage we aggregate global features into local features towards solving the \textit{ineffective} problem. Afterwards, a novel \textit{Long Range Memory} is introduced to solve the \textit{insufficient} problem. Experiments conducted on ImageNet VID dataset validate the effectiveness of our method. 

\section*{Acknowledgement}

This work is supported by National Key R\&D Program of China (2018YFB1402600), BJNSF (L172037), Beijing Acedemy of Artificial Intelligence and Zhejiang Lab.

\newpage
{\small
\bibliographystyle{ieee_fullname}
\bibliography{reference}
}

\end{document}